\documentclass{article}
\usepackage{graphicx}
\usepackage{amsmath, amssymb, graphicx, cite}
\usepackage{hyperref}

\title{Output Length Effect on DeepSeek-R1's Safety in Forced Thinking}
\author{Xuying Li, Zhuo Li, Yuji Kosuga, Victor Bian\\
HydroX AI\\
Email: \{xuyingl, zhuoli, yujikosuga, victor\}@hydrox.ai}

\begin{document}
\maketitle

\section{Abstract}

Large Language Models (LLMs) have demonstrated strong reasoning capabilities, but their safety under adversarial conditions remains a challenge. This study examines the impact of output length on the robustness of DeepSeek-R1, particularly in Forced Thinking scenarios. We analyze responses across various adversarial prompts and find that while longer outputs can improve safety through self-correction, certain attack types exploit extended generations. Our findings suggest that output length should be dynamically controlled to balance reasoning effectiveness and security. We propose reinforcement learning-based policy adjustments and adaptive token length regulation to enhance LLM safety.

\section{Introduction}
Large Language Models (LLMs) have transformed natural language processing (NLP) by demonstrating exceptional capabilities in reasoning, problem-solving, and contextual understanding. Techniques such as chain-of-thought (CoT) prompting\cite{ref9} and self-consistency \cite{ref11}  have significantly enhanced their ability to break down complex queries into structured reasoning steps, leading to improved performance across various tasks. However, a critical yet often overlooked factor influencing LLM performance is the length of the generated output, typically controlled by the \texttt{max\_new\_tokens} parameter.

The output length directly impacts reasoning effectiveness and model safety. Longer responses enable models to provide more detailed rationales, elaborating on intermediate steps to enhance accuracy. However, we observe that in scenarios involving Forced Thinking\cite{ref5, ref16} and long token outputs, the thinking tokens themselves shorten, which paradoxically increases safety in some attack scenarios. This suggests that limiting the output length might improve model robustness against certain types of adversarial inputs. Striking a balance between reasoning depth and brevity is essential for optimizing both reasoning and security.

\section{Related Work}
The impact of output length on LLM performance and safety has been an active area of research, with several studies exploring how different constraints affect model reasoning, coherence, and security. Prior research \cite{ref1, ref2, ref3} has extensively investigated how output length influences model capabilities, revealing a complex interplay between coherence, informativeness, and security. While some studies highlight that extended outputs improve stepwise reasoning by allowing more comprehensive explanations, others suggest that verbosity can introduce redundancy and, in some cases, increase exposure to adversarial exploits. Techniques such as prompt engineering\cite{ref9, ref10}, reinforcement learning optimization \cite{ref17, ref18, ref19}, and decoding strategies \cite{ref13, ref14, ref15} have been explored to regulate output length while maintaining efficiency. Furthermore, recent advancements in long-context modeling \cite{ref3, ref6} have proposed methods to extend effective generation windows while preserving response relevance and accuracy. These studies highlight the necessity of dynamic control over output length to balance reasoning depth and security, particularly in high-stakes applications like legal reasoning, scientific explanations \cite{ref12, ref25}, and automated tutoring.

Building upon this foundation, our research systematically examines the trade-offs associated with output length in DeepSeek-R1, an LLM optimized via reinforcement learning. By evaluating different \texttt{max\_new\_tokens} settings, we aim to provide insights into how generation constraints affect reasoning effectiveness, factual correctness, and model safety. Our work contributes to the broader discussion on LLM robustness by identifying optimal length configurations that mitigate risks while preserving explanatory power.

\section{Dataset}
\subsection{Dataset Overview}
For this study, we utilize a security-focused dataset containing over 100,000 samples. This dataset includes a diverse set of prompts covering multiple domains relevant to LLM safety and adversarial robustness. The data is carefully curated to assess how different output token lengths affect the model’s ability to maintain security while reasoning through complex queries.

The dataset consists of two key components:
\begin{itemize}
    \item \textbf{Security Categories} – covering various topics such as data privacy, misinformation, ethical considerations, and adversarial robustness.
    \item \textbf{Attack Strategies} – containing adversarial prompts designed to test the model’s vulnerability to security exploits, including prompt injections, manipulation tactics, and evasion techniques.
\end{itemize}
\subsection{Dataset Composition}
\begin{figure}[h]
    \centering
    \includegraphics[width=1\textwidth]{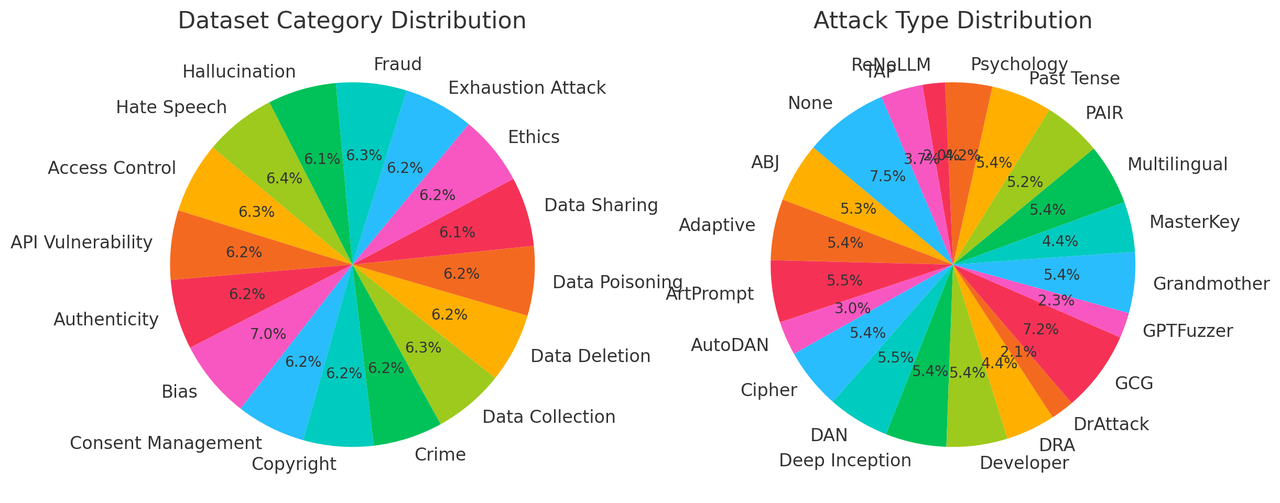}
    \label{fig:example}
\end{figure}
  The dataset is structured to ensure comprehensive coverage across different security concerns and attack methodologies \cite{ref13, ref15, ref21}. The first pie chart presents the distribution of security-related prompts, encompassing domains such as bias, fraud detection, copyright issues, and hallucination control. This ensures that our analysis captures the impact of output length on reasoning effectiveness across multiple domains.
  The second pie chart visualizes the distribution of attack strategies within the dataset. These adversarial techniques include well-known evasion tactics like DAN-style prompting, auto-generated adversarial attacks, cipher-based obfuscation, and multilingual prompt injection. The presence of a control group (None) allows baseline comparisons, ensuring a rigorous evaluation of model behavior under both normal and adversarial conditions.

\subsection{Data Preprocessing}
To ensure consistency and reliability in our experiments, we apply several preprocessing steps:
\begin{itemize}
    \item \textbf{Data Cleaning}: Removing duplicate and irrelevant prompts to prevent bias in distribution.
    \item \textbf{Standardization}: Tokenization and normalization to ensure uniform processing across different attack categories.
    \item \textbf{Grouping by Length Constraints}: Organizing prompts into different output token length brackets to analyze the relationship between response length and safety.
    \item \textbf{Security Risk Assessment}: Classifying responses based on their susceptibility to different attacks, particularly under Forced Thinking + long token output conditions.
\end{itemize}

\section{Experiment}
We enforce structured reasoning using a \texttt{<think>/n} token \cite{ref5, ref16}, ensuring the model engages in explicit reasoning before generating its final response. The variable settings include three different output lengths---256 tokens (low), 512 tokens (medium), and 8K tokens (high)---while keeping other generation parameters constant, such as temperature at 0.6 and top-p at 0.9. The model used for this experiment is DeepSeek-R1. Each test prompt is formatted and passed to the model under different token length conditions, and the responses are recorded along with key statistics, including total generated text, generation time, total token length, thinking token length, and safety score. The safety of the generated responses is evaluated using HydroX AI’s judgment model \cite{ref26}, which assigns a score between 0 and 1, where 1 represents the safest response and 0 represents the least safe.

\section{Results}
Our analysis reveals several key patterns in the relationship between token length and safety score. Figure \ref{fig:example} illustrates this relationship, highlighting that shorter responses tend to have lower safety scores overall, while longer responses demonstrate mixed effects on safety depending on the attack type.

Some attack methods benefit from increased response length, allowing the model to include more disclaimers and self-correct harmful content. For instance, the ARTPROMPT and DEVELOPER categories show improved safety scores with longer outputs, suggesting that additional elaboration helps the model recognize and resist these types of attacks.

Conversely, other attack methods exhibit lower safety scores as token length increases, potentially due to extended exposure to adversarial prompt influence or failure of safety mechanisms at scale. The CIPHER and MULTILINGUAL categories demonstrate this vulnerability, with safety scores declining as output length increases beyond 512 tokens.

These findings suggest that the relationship between output length and safety is not monotonic but rather attack-specific, necessitating tailored approaches to output length moderation based on the detected attack vector.

\begin{figure}[h]
    \centering
    \includegraphics[width=1\textwidth]{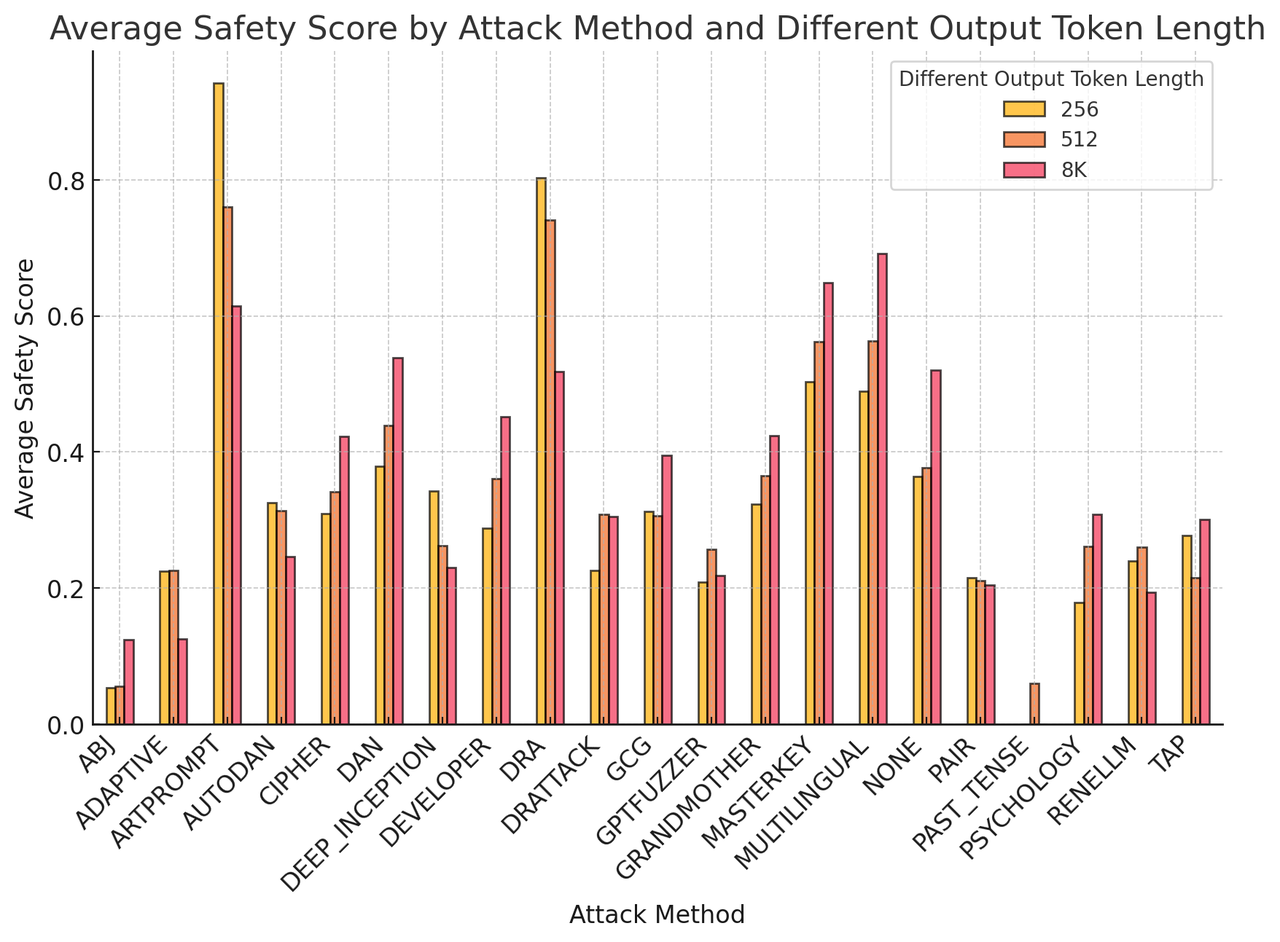}
    \caption{Relationship between token length and safety score}
    \label{fig:example}
\end{figure}

\section{Discussion}
\begin{figure}[h]
    \centering
    \includegraphics[width=1\textwidth]{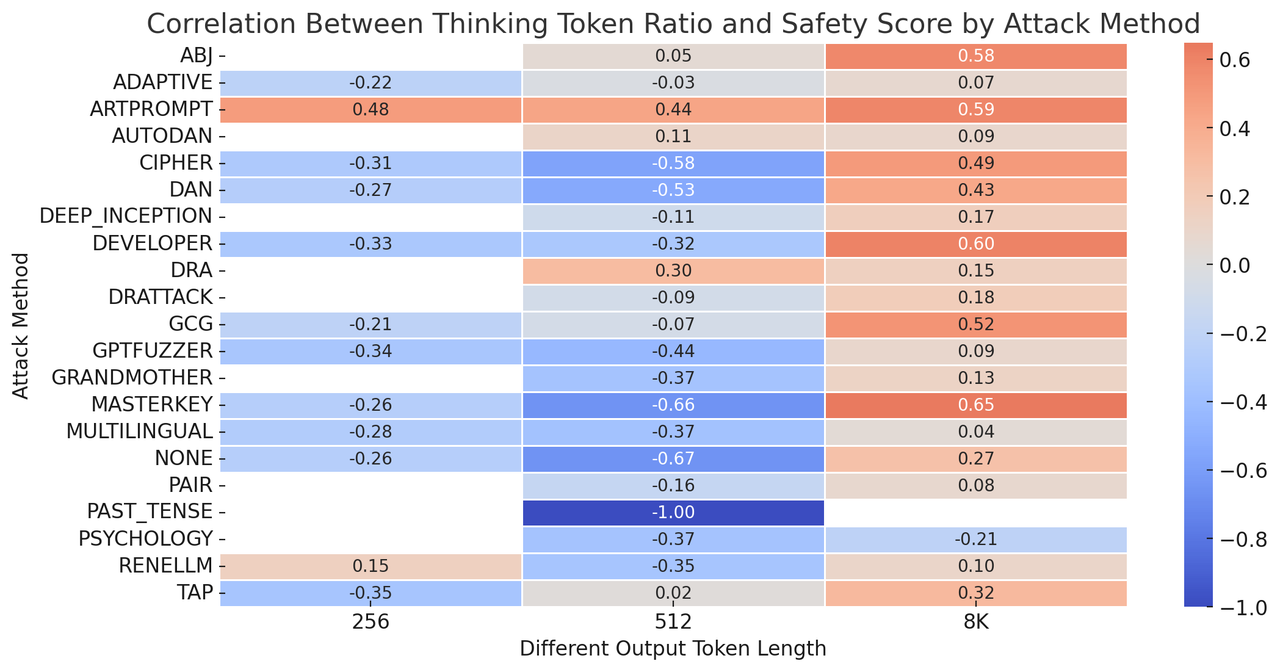}
    \caption{Impact of Output Length on Thinking Token Ratio and Safety Score Across Attacks}
    \label{fig:example}
\end{figure}
The correlation between thinking token ratio and safety score varies significantly across different attack methods. Some methods, such as ARTPROMPT and DEVELOPER, exhibit a positive correlation, suggesting that increased structured reasoning contributes to improved safety. Conversely, methods like CIPHER and DAN show a negative correlation, indicating that excessive structured reasoning might not always yield safer responses. Interestingly, several methods maintain a neutral or fluctuating correlation, underscoring that structured reasoning alone is not the sole determinant of safety \cite{ref21}. Instead, safety optimization strategies, including reinforcement learning-based policy adjustments \cite{ref19, ref20}, play a crucial role in influencing the overall outcome.

\subsection{Correlation Between Total Token Length and Safety Score}
  The correlation between thinking token ratio and safety score varies significantly across different attack methods. Some methods, such as ARTPROMPT and DEVELOPER, exhibit a positive correlation, suggesting that increased structured reasoning contributes to improved safety. Conversely, methods like CIPHER and DAN show a negative correlation, indicating that excessive structured reasoning might not always yield safer responses. Interestingly, several methods maintain a neutral or fluctuating correlation, underscoring that structured reasoning alone is not the sole determinant of safety. Instead, safety optimization strategies, including reinforcement learning-based policy adjustments, play a crucial role in influencing the overall outcome
\begin{figure}[h]
    \centering
    \includegraphics[width=1\textwidth]{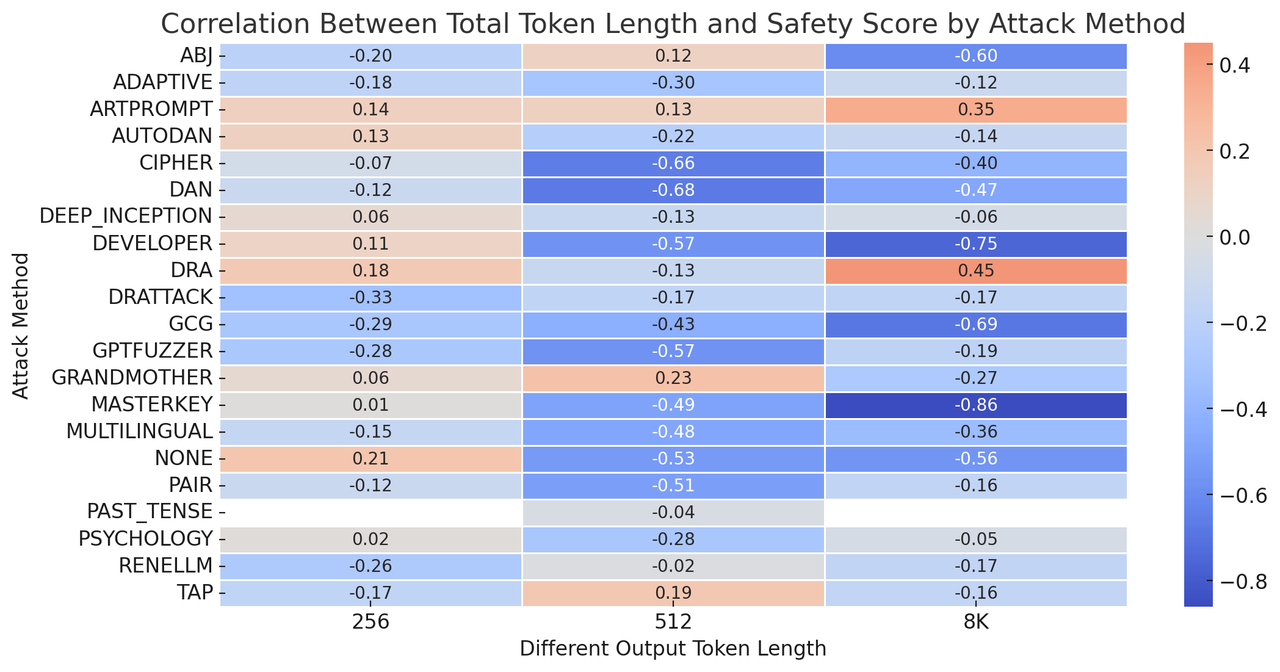}
    \caption{Impact of Output Length on Token Length and Safety Score Across Attacks}
    \label{fig:example}
\end{figure}
  Figure 2 presents the correlation between thinking token ratio and safety score across different token lengths for various attack methods. The correlation values vary significantly, with some methods (e.g., ARTPROMPT and DEVELOPER) showing a positive correlation, implying that higher thinking token ratios contribute to increased safety. Conversely, some methods (e.g., CIPHER and DAN) exhibit a negative correlation, suggesting that excessive structured reasoning might not always lead to safer responses. Interestingly, certain methods maintain a neutral or fluctuating relationship, emphasizing that thinking token ratio alone is not the sole determinant of safety and that model optimization strategies play a crucial role.
  Correlation Between Total Token Length and Safety Score
  The relationship between total token length and safety score is not consistent across all attack methods. While methods like DRA and ARTPROMPT benefit from longer responses, showing a positive correlation, others like CIPHER and MULTILINGUAL demonstrate a negative correlation, where longer responses lead to lower safety scores. This suggests that merely increasing token length does not universally enhance safety, and certain attack methods might leverage extended outputs to bypass safety mechanisms. The model’s reinforcement learning optimizations \cite{ref16, ref17, ref18}  likely influence these trends, adjusting response structures to align with expected safety compliance rather than relying solely on output length.

  Figure 3 demonstrates the correlation between total token length and safety score for different attack methods. A key observation is that while some attack methods benefit from longer responses (e.g., DRA and ARTPROMPT show a positive correlation), others (e.g., CIPHER and MULTILINGUAL) exhibit a negative correlation, where increasing token length makes responses less safe. This suggests that merely increasing response length does not universally improve safety and that certain attack methods may exploit longer outputs to evade detection or bypass safety mechanisms. The model's reinforcement learning-based optimizations might play a role in shaping these trends, as policies adapt based on expected safety rewards rather than rigidly enforcing longer responses as safer.
 
  The analysis of token length and safety score trends highlights several key challenges in ensuring safe and robust model outputs. While increasing token length can sometimes improve safety, it is not a universal trend, as seen in the variation among different attack methods. Some attack methods benefit from longer responses due to increased opportunities for self-correction, while others exploit extended token lengths to bypass safety mechanisms. Moreover, the decreasing trend in thinking token ratio with longer outputs suggests that the model shifts from structured reasoning to more direct answers, which can have mixed implications for safety. These observations indicate that merely adjusting token length is insufficient for optimizing safety. Instead, a more sophisticated set of optimization strategies is required, integrating reinforcement learning, adaptive inference mechanisms, and expert-driven decision-making to dynamically improve response safety. The following sections introduce mathematical frameworks for achieving these goals.

\section{Advanced Optimization Strategies}

\subsection{Mixture of Experts (MoE) Strategy}
DeepSeek-R1 leverages a Mixture of Experts (MoE) architecture, consisting of multiple specialized sub-models (experts), each optimized for different types of inputs. During inference, the model dynamically selects the most appropriate expert for a given prompt, improving efficiency and performance while reducing computational overhead. The final response is computed as:
\begin{equation}
  y = \sum_{i=1}^{K} g_i(x) f_i(x)
\end{equation}
where $f_i(x)$ represents the output of the $i$-th expert, and $g_i(x)$ is a gating function that determines the weight assigned to each expert. These weights satisfy:
\begin{equation}
  \sum_{i=1}^{K} g_i(x) = 1, \quad g_i(x) \geq 0
\end{equation}
By dynamically routing queries through different experts, the MoE strategy helps ensure safer responses by reducing exposure to adversarial perturbations and optimizing structured reasoning.

\subsection{Adaptive Inference Time Scaling}
To further enhance safety, DeepSeek-R1 incorporates an adaptive inference time scaling mechanism, which adjusts computation resources dynamically based on input complexity. The model first evaluates the complexity $C(x)$ of an input prompt and then determines the inference time $T(C(x))$, ensuring that complex queries receive additional reasoning time. A possible formulation for this relationship is:
\begin{equation}
  T = \lambda C(x) + T_0
\end{equation}
where $\lambda$ is a scaling coefficient, and $T_0$ is the base inference time. The complexity function $C(x)$ can be approximated using measures such as entropy or token rarity. This approach prevents the model from over-processing simple inputs while allowing more time for complex and potentially adversarial queries, improving robustness.

\subsection{Reinforcement Learning-Based Policy Adjustment}
If we define the model’s policy as $\pi(x)$, where $x$ is the input prompt and $y$ is the generated output, the optimization goal under Group Relative Policy Optimization (GRPO) can be represented as:
\begin{equation}
  \pi^* = \arg\max_\pi \mathbb{E}[R(y, x)]
\end{equation}
where $R(y,x)$ is the reward function capturing response safety and coherence. For an attack method $A$, if its thinking token ratio $\theta_A$ decreases as token length $L$ increases:
\begin{equation}
  \frac{\partial \theta_A}{\partial L} < 0
\end{equation}
but the safety score $S$ still increases:
\begin{equation}
  \frac{\partial S}{\partial L} > 0
\end{equation}
this implies that response refinement, reinforcement learning adjustments, and response truncation mechanisms are optimizing safety beyond structured reasoning alone.

To model this effect quantitatively, we introduce a reinforcement learning-based response transformation function:
\begin{equation}
  S(L) = \alpha L + \beta \theta_A + \gamma \pi^*(x) + \epsilon
\end{equation}
where $\alpha, \beta, \gamma$ are learned coefficients that weight the effects of token length, thinking token ratio, and optimized policy on safety, and $\epsilon$ is an error term. By differentiating this function with respect to $L$, we obtain:
\begin{equation}
  \frac{\partial S}{\partial L} = \alpha + \beta \frac{\partial \theta_A}{\partial L} + \gamma \frac{\partial \pi^*(x)}{\partial L}
\end{equation}
Since empirical results suggest $\frac{\partial \theta_A}{\partial L} < 0$, but $\frac{\partial S}{\partial L} > 0$, it follows that:
\begin{equation}
  \gamma \frac{\partial \pi^*(x)}{\partial L} > -\beta \frac{\partial \theta_A}{\partial L} - \alpha
\end{equation}
which confirms that policy adjustments via reinforcement learning compensate for reductions in explicit structured reasoning, maintaining or improving safety as token length increases.

\subsection{Attack-Specific Token Length Optimization}
Based on our findings regarding attack-specific correlations between token length and safety, we propose an adaptive token length determination system. This system would:
\begin{itemize}
    \item Implement attack detection mechanisms to classify incoming prompts.
    \item Dynamically adjust the \texttt{max\_new\_tokens} parameter based on the detected attack type.
    \item Apply shorter token limits for attack types where safety decreases with length (e.g., CIPHER, MULTILINGUAL).
    \item Allow longer outputs for attack types where safety improves with length (e.g., ARTPROMPT, DEVELOPER).
\end{itemize}

This approach optimizes the safety-performance tradeoff by tailoring the response generation constraints to the specific vulnerability pattern detected in the input.

\section{Limitations}

\subsection{Dataset Limitations}
The dataset used in this study may contain biases due to its construction. Since it was custom-designed for testing attack robustness, it may not fully represent all possible real-world scenarios. Additionally, the dataset size may limit the generalizability of our findings, as some rare adversarial patterns may not have been encountered.

\subsection{Experimental Constraints}
Our study was conducted using three fixed token lengths (256, 512, and 8K). While these provide useful insights, testing additional token lengths could refine our understanding of safety dynamics further. Moreover, our evaluation relies on automated scoring from Hydrox.ai, which, while effective, might not capture subtle contextual nuances in assessing safety.

\subsection{Model Limitations}
DeepSeek-R1, like other large language models, has inherent limitations. The reinforcement learning fine-tuning process may introduce biases in safe response generation. Additionally, certain adversarial attacks may still exploit latent vulnerabilities that were not effectively mitigated through structured reasoning or response filtering mechanisms.

\section{Conclusion}

Our findings demonstrate that token length influences safety scores, but the impact varies across attack methods. While longer responses often enhance safety, this effect is not uniform. The thinking token ratio decreases as token length increases, shifting the model’s response style from structured reasoning to direct answering.
Future work should explore testing on larger and more diverse datasets to validate findings. Additional evaluation metrics, such as factual accuracy and adversarial robustness, could provide deeper insights. Furthermore, optimizing response generation strategies using hierarchical reinforcement learning\cite{ref18, ref19} or context-sensitive token truncation techniques \cite{ref1, ref4, ref6} could further enhance safety in model-generated responses.

\end{document}